\documentclass[letterpaper, 10 pt, conference]{ieeeconf}

\IEEEoverridecommandlockouts                              

\usepackage[l2tabu,orthodox]{nag}

\usepackage[
    backend=bibtex,
    bibencoding=utf8,
    style=ieee,
    sorting=none,
    natbib=true,
    doi=false,
    isbn=false,
    url=false,
    eprint=false,
    maxcitenames=1,
    mincitenames=1,
]{biblatex}

\usepackage[draft,pdftex,colorlinks]{hyperref}

\usepackage[printonlyused]{acronym}

\usepackage{siunitx}
\usepackage[all]{nowidow}

\usepackage[dvipsnames]{xcolor}

\usepackage{lipsum}

\usepackage[pdftex]{graphicx}

\usepackage{epstopdf}

\usepackage{import}

\graphicspath{{./latexGoodPractices/}}

\usepackage{booktabs}

\usepackage{tabu}
\usepackage{makecell}
\usepackage{float}

\usepackage{amssymb,amsfonts,amsmath,amscd}

\usepackage{cleveref}
\creflabelformat{equation}{#2#1#3}

\usepackage{bm}

\newcommand{\bbm}{\begin{bmatrix}}
        \newcommand{\ebm}{\end{bmatrix}}

\usepackage[normalem]{ulem} 

\usepackage{microtype} 

\makeatletter
\usepackage[utf8]{inputenc}
\usepackage{commath}
\usepackage{mathtools}
\usepackage{mathabx}
\usepackage{siunitx}
\usepackage[ruled,vlined]{algorithm2e}
\usepackage[lofdepth,lotdepth]{subfig}
\usepackage{tikz}
\usepackage{tabularray}
\usepackage{physics}

\let\oldtheequation\theequation
\renewcommand\tagform@[1]{\maketag@@@{\ignorespaces#1\unskip\@@italiccorr}}
\renewcommand\theequation{(\oldtheequation)}
\makeatother

\usepackage[belowskip=0pt,aboveskip=1pt]{caption}
\usepackage[normalem]{ulem}

\newcommand{\ie}{i.e., }

\newcommand{\bT}{\bm{T}}

\newcommand{\bC}{\bm{C}}

\newcommand{\bomega}{\bm{\omega}}

\newcommand{\btau}{\bm{\tau}}

\newcommand{\br}{\bm{r}}
\newcommand{\ba}{\bm{a}}

\newcommand{\bI}{\bm{I}}

\newcommand{\R}{\mathbb{R}}

\DeclareMathOperator*{\argmin}{arg\,min}

\usepackage{amssymb}
\usepackage{pifont}

\acrodef{ICP}{Iterative Closest Point}
\acrodef{DOF}{Degree Of Freedom}
\acrodef{INS}{Inertial Navigation System}
\acrodef{GNSS}{Global Navigation Satellite System}
\acrodef{GPS}{Global Positioning System}
\acrodef{UGV}{Unmanned Ground Vehicle}
\acrodef{UAV}{Unmanned Aerial Vehicle}
\acrodef{MAV}{Micro Aerial Vehicle}
\acrodef{IMU}{Inertial Measurement Unit}
\acrodef{SLAM}{Simultaneous Localization and Mapping}
\acrodef{MEMS}{Micro-Electromechanical Systems}
\acrodef{GTSAM}{Georgia Tech Smoothing and Mapping Library}
\acrodef{LeGO-LOAM}{Lightweight and Ground Optimized Lidar Odometry and Mapping}
\acrodef{KF}{Kalman Filter}
\acrodef{RPE}{Relative Pose Error}

\DeclareUnicodeCharacter{0308}{¨} 

\begin{document}
\title{\LARGE \textbf{Under Pressure: Altimeter-Aided ICP for 3D Maps Consistency}}

\author{William Dubois,$^{1}$ Nicolas Samson,$^{1}$ Effie Daum,$^{1}$ Johann Laconte$^2$ and François Pomerleau$^{1}$
  \thanks{$^{1}$Northern Robotics Laboratory, Université Laval, Québec City, Canada,
    {\texttt{\small{$\{$william.dubois, nicolas.samson, effie.daum, francois.pomerleau$\}$ @norlab.ulaval.ca}}}}%
  \thanks{$^{2}$Université Clermont Auvergne, INRAE, UR TSCF, 63000, Clermont-
Ferrand, France {\texttt{\small{johann.laconte@inrae.fr}}}}%
}

\linepenalty=3000
\addtolength{\textfloatsep}{-0.1in}

\maketitle
\thispagestyle{empty}
\pagestyle{empty}


\begin{abstract}

We propose a novel method to enhance the accuracy of the \ac{ICP} algorithm by integrating altitude constraints from a barometric pressure sensor.
While \ac{ICP} is widely used in mobile robotics for \ac{SLAM}, it is susceptible to drift, especially in underconstrained environments such as vertical shafts.
To address this issue, we propose to augment \ac{ICP} with altimeter measurements, reliably constraining drifts along the gravity vector.
To demonstrate the potential of altimetry in \ac{SLAM}, we offer an analysis of calibration procedures and noise sensitivity of various pressure sensors, improving measurements to centimeter-level accuracy. 
Leveraging this accuracy, we propose a novel \ac{ICP} formulation that integrates altitude measurements along the gravity vector, thus simplifying the optimization problem to 3\nobreakdashes-\ac{DOF}.
Experimental results from real-world deployments demonstrate that our method reduces vertical drift by \textbf{84\%} and improves overall localization accuracy compared to state-of-the-art methods in non-planar environments.
\end{abstract}



\acresetall
\section{Introduction}

\begin{figure}[t]
\centering
\includegraphics[width=\linewidth]{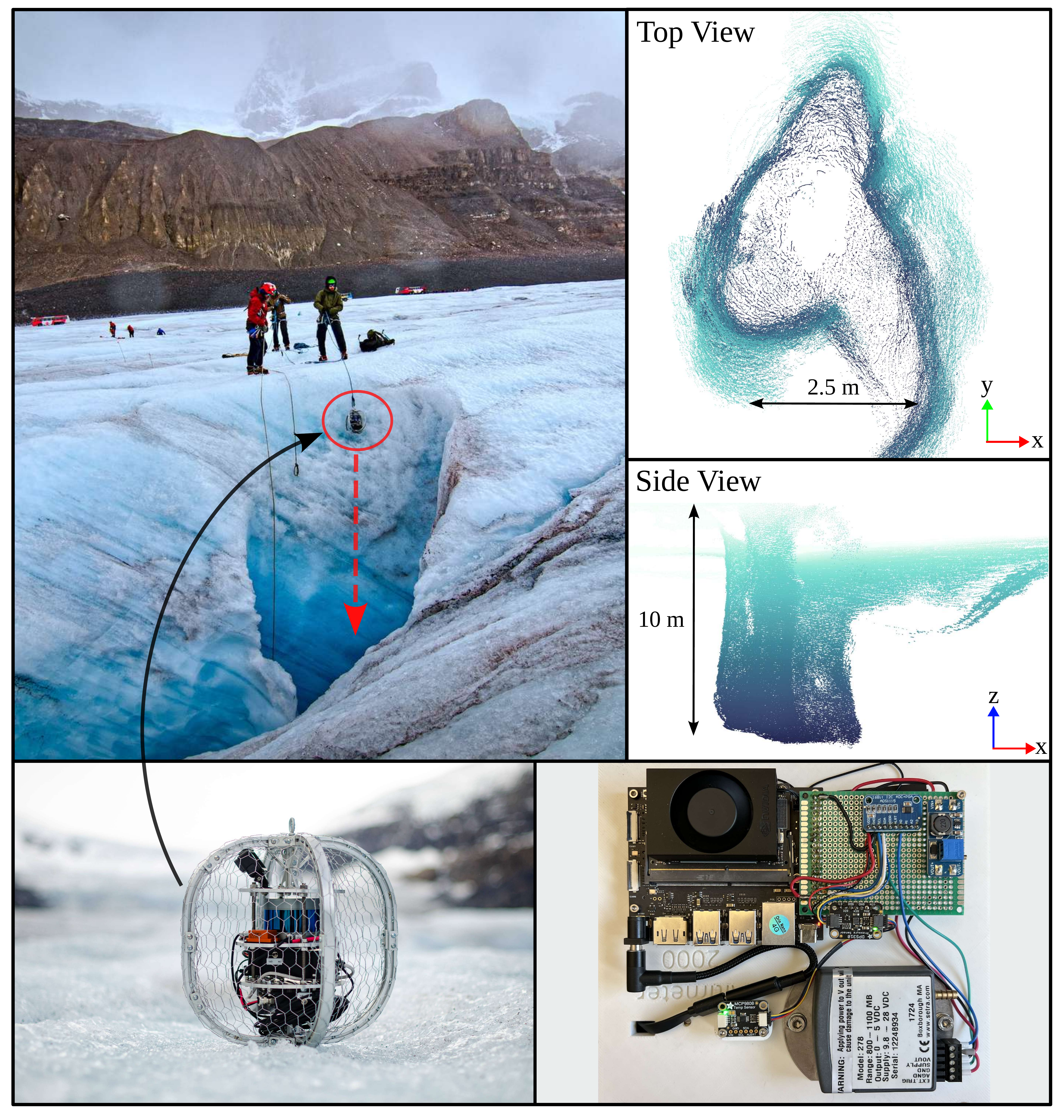}
\caption{\emph{Top Left}: A robotic platform deployed into a glacier moulin, a naturally occurring crevasse formed by water infiltration within the ice with low vertical constraints. 
\emph{Top Right}: a 3D map of the moulin, illustrating its structure and dimensions. 
\emph{Bottom Left}: A custom-built sensing platform for lidar-\ac{SLAM}. 
\emph{Bottom Right}: The altimeter rig used for sensor calibration and as a reference station throughout our experiments.}
\label{fig:fig-intro}
\end{figure}

Recently, mobile robots have gained increasing popularity in applications such as inspection, environmental surveying, and search and rescue tasks.
To efficiently accomplish those missions, the lidar sensor is the most popular choice for accurate localization \citep{Ebadi2023}.
Although strong performances are attained with lidar-based \ac{SLAM} solutions \citep{Baril2022}, degenerate environments exist, such as corridors, glacier moulins, or long forest trails where constraints are insufficient and thus lead the localization to drift along under-constrained axes \citep{Kubelka2022}.
\autoref{fig:fig-intro} depicts a deployment carried out on the Athabasca Glacier \citep{dubois2024workshop}, where glacier moulins were investigated for environmental survey purposes. 
Due to the degenerate nature of the environment, we observed a significant drift along the vertical axis, causing poor performances of the \ac{ICP} algorithm.
These performances led to weaker performances in 3D environment reconstruction.
To effectively reduce this vertical drift, global information has been fused to lidar-based solutions, such as the gravity vector, derived from \ac{IMU} sensors \citep{Kubelka2022}, or \ac{GNSS} data \citep{Babin2019}.
However, \ac{GNSS} solutions have their limitations, as they will not work in situations when satellites are obstructed, and they also tend to be inaccurate in measuring altitude.
Moreover, relying solely on the gravity vector will still result in drift during extended deployments.

A viable option for constraining a \ac{DOF} and limiting drift in \ac{SLAM} solutions along the vertical axis is incorporating barometric altimetry.
This method calculates an altitude value, known as pressure altitude, from a barometric pressure measurement \citep{Hipkin2000}. 
The \ac{ICP} algorithm could thus benefit from more information by incorporating known constraints about the vertical axis.
However, challenges remain to enable centimeter-level accuracy localization with lidar-based \ac{SLAM} systems.
Factors like atmospheric conditions will induce high biases in measurements, leading to high uncertainty in the computation of absolute pressure altitude \citep{Bolanakis2017}.
Without extensive and precise access to meteorological data, using a base station as a reference is the only way to effectively reduce this atmospheric bias \citep{Bolanakis2017}.

In this paper, we propose a novel integration method of barometric pressure measurements, adding constraints to the \ac{ICP} optimization along the often under-constrained vertical axis, effectively reducing drift and improving \ac{SLAM} performances.
Specifically, our contributions are 1) A comprehensive bias compensation of barometric pressure measurements; 2) A characterization of the performances of ceramic pressure sensors and \ac{MEMS} pressure sensors in the context of localization; and 3) A novel tightly-coupled \ac{IMU}-barometer-lidar localization and mapping framework.

\section{Related work}
\label{sec:related_work}
Absolute altitude, mostly used in aeronautics, refers to the vertical distance above sea level. 
There are several types of altitude measurements, such as geometric, geopotential, and pressure altitude, respectively computed using the distance above the Earth's surface, the gravitational potential, and the atmospheric pressure \citep{Hipkin2000}. 
As geometric and geopotential altitudes necessitate a known reference point to reach an altitude in a global Earth frame, we steer away from these methods, given that most lidar-based localization systems use local reference frames specific to each robotic deployment.
Relative altitude measurement can be achieved by reading pressure from a static altitude and comparing it to a reading moving vertically.
Thus, our investigation focuses on relative altitude, with barometers readily available, highlighting key steps to increase the accuracy of these sensors to the centimeter level.

Barometric sensors can be manufactured following different processes, two common ones relying on \ac{MEMS} or ceramic.
\ac{MEMS} barometric sensors are cheaper and widely used in the literature for positioning but are known to suffer from internal temperature bias.
\citet{Bolanakis2017} used a \ac{MEMS}-based sensor in an absolute and a relative altitude framework, highlighting three important sources of bias when working with variations of minimally \SI{6}{\pascal}. 
The first bias comes from changes in ambient conditions and atmospheric conditions.
These changes would affect two sensors roughly in the same weather and thus can be mitigated by computing relative altitude.
The second bias is produced by the internal temperature variation of \ac{MEMS}, which needs calibration to be mitigated.
Finally, a constant bias offset caused by the nature of the \ac{MEMS} technology needs to be identified by comparing its reading to known values.
\citet{parviainen2008differential} conducted relative altitude experiments with a car using a \ac{MEMS} pressure sensor and identified air currents from wind and ventilation systems as the two main sources of error while driving with a resolution of \SI{10}{\pascal}. 
\citet{Sabatini2013} focused on short-time observations of \ac{MEMS} pressure sensor measurements, thus ignoring the drift caused by the sensor's internal temperature and atmospheric variations. 
They modeled the remaining correlated and uncorrelated measurement noise to obtain more accurate measurements with an uncertainty of \SI{0.1}{m}.
In contrast, \citet{Matyja2022} investigated and modeled three bias sources with a resolution of \SI{1}{\pascal}. 
First, the principal error caused by nonstandard sea level conditions was modeled using the difference between current and standard sea level conditions.
Second, external disturbances caused by temperature, vibrations, and air movements were estimated based on the velocity of air masses and their density.
Finally, the drift error was modeled as a white Gaussian stochastic process as the uncorrelated noise by \citet{Sabatini2013}.
In robotic deployments, the aim is to conduct experiments spanning from hours to days. 
Therefore temperature calibration and compensation must be addressed for \ac{MEMS} sensors as the assumption of short-time measurement does not hold.
The desired resolution is also \SI{0.1}{\pascal}, which has not been reached in the literature, to achieve the accuracy level necessary for localization.
In this paper, we leverage the work of \citet{Bolanakis2017} to mitigate bias by relying on one static and one mobile sensor.
Moreover, we investigate the performance of recent ceramic-based sensors in the context of robotic deployments and \ac{SLAM}.
To the best of our knowledge, this is the first work evaluating the performances of ceramic-based barometric sensors in the context of \ac{SLAM}, as they are aimed toward meteorological use.

The fusion of barometer measurements to localization, odometry, and \ac{SLAM} solutions has already proven promising results.
\citet{Zaliva2014} improved the \ac{GPS} altitude estimation using a barometric pressure sensor.
They took advantage of the rate of pressure sensor measurements, as opposed to the intermittent errors caused by satellite obstruction of \ac{GPS}, and computed the absolute altitude.
Also, they demonstrated that using \ac{GPS} measurements to estimate the absolute altitude drift allows to correct the pressure altitude measurement and reach a confidence bound \SI{85}{\percent} smaller than that of the typical \ac{GPS} altitude estimation.
However, in robotic deployments, the confidence bound attained of about \SI{1}{\metre} is far from reaching the centimeter-level accuracy needed to support lidar-based localization.
\citet{Urzua2019} introduced barometer measurements as vertical constraints to their \ac{UAV} system to update the altitude of their vehicle.
To do so, they compute an initial absolute altitude subtracted from every following altitude estimation.
They also counteract the drift of the absolute altitude by modeling the sensor bias as a Gauss-Markov process and assuming white Gaussian noise in the measurement.
Even though they try to model the drift, this approach does not reach the centimeter-accuracy level desired for lidar \ac{SLAM}, as they compute an absolute altitude which by definition drifts with varying atmospheric conditions.
\citet{Song2015} use a flight control unit that encapsulates barometer and inertial measurements, computing the attitude of their \ac{UAV} and augmenting the flight stability, but this method only reaches an accuracy of \SI{20}{\centi\metre} and is not applicable for the robotic deployments aimed for in this work.
\citet{Talbot2023} fuse a \ac{MEMS}-based barometer with other sensors through the barometric factor introduced in \ac{GTSAM}\citep{Dellaert2012}.
This factor uses the measurement and a noise model estimated as a zero-mean Gaussian.
As shown with the atmospheric conditions bias source, the zero-mean Gaussian assumption does not hold as the measurement will drift with ambient changes and not only suffer from white noise \citep{Bolanakis2017}.
\citet{Leisiazar2022} also included a barometer in their lidar-based \ac{SLAM} algorithm to enable the use of elevators when mapping multi-floor buildings.
Their work relies on an elevator-door detector, allowing them to record the pressure at the entrance of the elevator and a second one during the exit.
They propose to stop the lidar-based localization when detecting the entrance to an elevator, record the pressure, and detect the exit of the elevator to compute a relative altitude, and finally restart the lidar-based localization.
Although this is functional in their specific use case, it is hard to identify when to disable the mapping in more generic robotic deployments, which can include slops or staircases.
Moreover, their solution assumes that each floor is planar.
In this work, we propose integrating the \emph{relative pressure altitude} within the \ac{ICP} algorithm to constrain the optimization along the gravity vector after correcting the measurements to account for their biases. 
As such, our 3\nobreakdashes-{DOF} \ac{ICP} algorithm is the first tightly-coupled implementation of \ac{ICP} with altitude measurements for robotic platforms in any unknown environment.

\section{Theory}
\label{sec:theory}
This section first provides an overview of the pressure measurement, its inherent biases, and how to mitigate them, then altitude computation is established based on the ideal gas law and dry air constant. Finally, our implementation of 3\nobreakdashes-\ac{DOF} minimization for the \ac{ICP} algorithm is formulated and expanded to highlight the simplification of the minimization problem.

\subsection{Altitude measurement}
\subsubsection{Pressure measurement}
\label{sec:theoryA}

Altitude measurements are computed from a sensor measuring atmospheric pressure $p$ (in \si{\pascal}) from the deformation of a membrane.
Barometric pressure sensors suffer from inherent noises and biases caused by the sensors themselves or external factors, one of them being the temperature of the sensor changing when powered.
High-grade barometric sensors rely on stable materials such as ceramic and gold at the expense of size, weight, and price.
Smaller \ac{MEMS} pressure sensors need to be calibrated for temperature variation as they dilate and contract during operations.
Calibration parameters $\bm{A}$ are usually provided by the manufacturer as a form of 2D polynomial function outputting the calibrated pressure $p_{\text{cal}}$ as a function of the raw pressure $p_{\text{raw}}$ and the temperature of the sensor $t$ (in \si{\degreeCelsius}).
A generic form of that compensation equation can be given as $p_{\text{cal}} = \bm{p}\bm{A}\bm{t}$ with 

\begin{equation} 
\begin{split}
    \bm{p} =& \begin{bmatrix} 
    \phantom{c_{00}}1 & p_\text{raw} & p_\text{raw}^2 & \dots& p_\text{raw}^N 
    \end{bmatrix}, 
    \\
    \bm{A} =& \begin{bmatrix}
        c_{00} & c_{01} & c_{02} & \dots & c_{0N}\\
        c_{10} & c_{11} & c_{12} &  & c_{1N}\\
        \vdots & & & \ddots & \vdots\\
        c_{N0}\phantom{^2} & c_{N1} & c_{N2} & \dots & c_{NN}
    \end{bmatrix},
    \begin{bmatrix}
    1\phantom{^2} \\ t\phantom{^2} \\ t^2 \\ \vdots \\ t^N
    \end{bmatrix} = \bm{t}, 
\end{split}
\label{equation_calibration}
\end{equation}

where both vectors $\bm{p}$ and $\bm{t}$ are representing measurements raised to the power zero to $N$.
It is often difficult to have details on the calibration procedure from a manufacturer.
Using a ceramic sensor as a reference, we record the pressures at different temperatures and minimize $p_\text{cal} = f(p_\text{raw}, t)$.
The residual error can estimate the variance $\sigma_p^2$ of the sensor.

It is also known that wind and air movements impact all barometric pressure sensors since they apply additional pressure to the sensor \citep{parviainen2008differential}.
To overcome this issue, enclosures can be designed with a leaking rate acting as a mechanical low-pass filter. 
Finally, pressure sensors are prone to measurement noise, which will differ depending on the technology used for the fabrication.
Within the context of supporting the localization of a vehicle, a numerical low-pass filter can be designed to reduce the noise level without adding delays inhibiting the motion dynamic of the robot. 
One should note that a \ac{KF} using a constant observation model with an observation noise equivalent to the calibration variance $\sigma_{p}^2$ is equivalent to a low-pass filter.

\subsubsection{Altitude computation}
To measure an altitude $z$ (in \si{\m}), pressure must be converted by making assumptions using the dry air gas constant $R_\text{dry}$ and a model on how the temperature $T(z)$ (in \si{\kelvin}) evolves with altitude. 
Note that we use $t$ for temperature in Celsius and $T$ for temperature in Kelvin.
The relation between two pressures $p_0 > p_1$ and two altitude $z_0 < z_1$ is given by the \textit{generic hypsometric equation}, formulated as
\begin{equation}
    \ln{\left(\frac{p_1}{p_0}\right)} = -\frac{g}{R_\text{dry}} \int_{z_0}^{z_1}\frac{dz}{T(z)},
    \label{equation_hyps_generic}
\end{equation}
where $g$ is the gravity.
A common simplification of this model is to assume that the temperature profile between $z_0$ and $z_1$ can be estimated with a virtually constant temperature $\overline{T}_v$ being the average of both layers~\citep{wallace2006atmospheric}.
With this assumption, the integral of \autoref{equation_hyps_generic} can be simplified to compute directly the difference of altitude $\delta z$ of two pressure sensors using
\begin{equation}
    \delta z = z_1 -z_0 = \frac{R_\text{dry}\overline{T}_v}{g} \ln{\left(\frac{p_0}{p_1}\right)}.
    \label{equation_alt_hyps}
\end{equation}
For a localization system of a ground vehicle, using two pressure sensors, one static and one on the vehicle, has the advantage of mitigating the impact of atmospheric phenomena as both sensors will drift in the same direction following the weather.
Moreover, the variation in altitude $\delta z$ is so small compared to aviation applications that the virtual temperature $\overline{T}_v$ can be assumed to be the same for both sensors and represent the average air temperature of the environment.

A second formulation, widely used in the literature \citep{Bolanakis2017, parviainen2008differential, Sabatini2013, Matyja2022}, is the \textit{barometric formula}.
This model approximates the variation of temperature between the sea level and a given altitude, assuming a temperature lapse rate  $\Gamma$.
For low altitudes between zero to \SI{11}{\km}, the U.S. Standard Atmosphere fixes this rate $\Gamma$ to a loss of \SI{0.0065}{\celsius} per meter.
For a ground vehicle state estimation with small $\delta z$, this model does not provide a better correction than the \textit{generic hypsometric equation} with average temperature.
In practice, both formulations lead to the same results, so in the remainder of this paper, variation in altitude is computed using the hypsometric equation of \autoref{equation_alt_hyps}.

\subsection{3\nobreakdashes-\ac{DOF} \ac{ICP} minimization}
The \ac{ICP} minimization process estimates a rigid transformation $\widehat{\mathbf{T}}$ between a reference set of 3D points $\mathcal{Q}$ (i.e., a map) and a set of 3D points  $\mathcal{P}$ (i.e., a scan point cloud) that have previously been pre-aligned using an initial pose estimate $\widecheck{\mathbf{T}}$.
In our case, $\widecheck{\mathbf{T}}$ is estimated by fusing information from the \ac{IMU}, wheel encoders, and the calculated altitude.
The process then minimizes an error function $e$, which has a minimum value if $\mathcal{P}$ and $\mathcal{Q}$ are well aligned.
Thus, to compute the optimal transformation $\widehat{\mathbf{T}}$ to register the set of points $\mathcal{P}$ in the set of points $\mathcal{Q}$, we have to solve
\begin{equation}
    \widehat{\mathbf{T}} = \argmin_{\bm{T}\,\in\,SE(3)} e(\mathcal{Q}, \mathcal{P}).
    \label{equation_6}
\end{equation}

Altitude measurements are aligned with the gravity vector.
Therefore, we base our approach on the gravity-constrained definition of point-to-plane error minimization introduced by \citet{Kubelka2022} and rotate point clouds to align with the gravity vector measured by an \ac{IMU}.
We extend their work to a 3\nobreakdashes-\ac{DOF} minimization by constraining the roll and pitch using the gravity vector and then $z$ coordinates using the variation in altitude.
First, we define the constrained transformation $\bT(\btau)=\left(\bC(\btau), \br(\btau)\right), \btau\in\R^3$ as 
\begin{equation}\\
\begin{aligned}
    \bC(\bm\tau) &= \exp\left(\gamma\bomega_z^\wedge\right) &\rightarrow& \quad \text{constrained rotation}\\
    \br(\bm\tau) &= \begin{bmatrix} r_x & r_y & 0 \end{bmatrix}^T  &\rightarrow& \quad \text{constrained translation}
    \label{equation_7}
\end{aligned}
\end{equation}
with $\bm\tau = \begin{bmatrix} \gamma & r_x & r_y \end{bmatrix}^T$ and where $\gamma$ represents the rotational angle around the $\mathbf{z}$ axis, $\bomega_z=\begin{bmatrix}0\quad 0\quad 1\end{bmatrix}^T$ and  $\bomega_z^\wedge\in\R^{3\times 3}$ is the associated skew-symmetric matrix \citep{Barfoot2017}.
As such, the resulting transformation is only able to translate in the $xy$ plane and perform a yaw rotation.
Then, we continue with the definition of the point-to-plane error function
\begin{equation}
    e(\mathcal{Q}, \mathcal{P}) = \sum_{k=1}^{K}\left(\mathbf{n}_k^T \begin{bmatrix}\bC(\bm\tau)\mathbf{p_k} + \br(\bm\tau)
     - \mathbf{q}_k\end{bmatrix}\right)^2,
    \label{equation_8}
\end{equation}
where $\bm{q_k}$ $\in$ $\mathbb{R}^3$, $\bm{p_k}$ $\in$ $\mathbb{R}^3$ are paired points from $\mathcal{Q}$ and $\mathcal{P}$ at index $k$, and $\mathbf{n_k}\in\mathbb{R}^3$ is the normal vector at point $\bm{q_k}$.
Following the minimization from \citet{Pomerleau2015}, adapted to the 3\nobreakdashes-\ac{DOF} implementation, the rotation matrix $\bC$ performs solely the yaw rotation. 
The other rotations are not necessary, as the point cloud $\mathcal{P}$ has been pre-aligned with the gravity, constraining the roll and pitch angles, according to the initial pose estimate $\widecheck{\mathbf{T}}$.

Using the small angle approximation $\exp(\gamma\bomega_z^\wedge)\approx \bI + \gamma\bomega_z^\wedge$, with $\bI\in\R^3$ being the identity matrix, \autoref{equation_8} can be simplified to
\begin{equation}
\begin{aligned}
    e(\mathcal{Q}, \mathcal{P}) &\approx \sum_{k=1}^{K}\left(\mathbf{n}_k^T\left[
        \left(\bI + \gamma\bomega_z^\wedge\right)\mathbf{p}_k + \br(\btau) - \mathbf{q}_k
    \right]\right)^2 \\
    &= \sum_{k=1}^{K}\left(\gamma\mathbf{n}_k^T\bomega_z^\wedge \bm{p}_k + \mathbf{n}_k^T\br(\btau) - \mathbf{n}_k^T\underbrace{(\mathbf{q}_k - \mathbf{p}_k)}_{\mathbf{d}_k}\right)^2 \\
    &=\sum_{k=1}^{K}\left(\ba_k^T\btau - \mathbf{n}_k^T\mathbf{d}_k\right)^2, \\
    &\quad\text{with } \ba_k = \begin{bmatrix}
        \mathbf{n}_k^T\bomega_z^\wedge \bm{p}_k & n^x_{k} & n^y_{k}
    \end{bmatrix}^T,
\end{aligned}
\label{equation_10}
\end{equation}
where $n^x_{k},n^y_{k}$ represents the x and y components of the normal vector $\bm{n}_k$, respectively.
From this, we minimize \autoref{equation_10} by setting its derivative to zero, as
\begin{equation}
\begin{aligned}
        &\dv{e}{\btau} = 2\sum_{k=1}^{K} \ba_k\left(\ba_k^T\btau-\mathbf{n}_k^T\mathbf{d}_k\right)
        =\bm{0} \\
        &\Leftrightarrow  \left(\sum_{k=1}^{K} \ba_k\ba_k^T\right)\btau = \sum_{k=1}^{K} (\mathbf{n}_k^T\mathbf{d}_k)\ba_k,
\label{equation_12}
\end{aligned}
\end{equation}
which can be solved using any linear solver.

\section{Results}
\label{sec:results}
This section first details the experimental setup used and the deployment environments where data was acquired to test and validate the proposed solution. 
Then, the \ac{MEMS} temperature calibration is presented, followed by the sensitivity analysis of both \ac{MEMS} and ceramic pressure sensors.
Finally, the localization and mapping performances of the proposed \ac{ICP} algorithm are compared to two baselines: the gravity constrained (i.e., 4\nobreakdashes-\ac{DOF} \citep{Kubelka2022}) and the regular \ac{ICP} algorithms (i.e., 6\nobreakdashes-\ac{DOF}~\citep{Pomerleau2013}).

\subsection{Experimental setup}
The platforms used throughout our experiments were a SuperDroid HD2 tracked robot, a custom-built sensing platform called the Sphere, and an altimeter sensor rig.
Both the Sphere and the altimeter rig are depicted in \autoref{fig:fig-intro}.
The altimeter rig was used to calibrate several DPS310 \ac{MEMS} sensors weighing \SI{1.5}{g} using a ceramic-based barometric pressure transducer: the Setra 278 weighing \SI{135}{g}.
Highlighting that in low-payload robotic deployments such as \acp{UAV}, ceramic-based sensors are not a viable option solely because of their weight.
The altimeter rig also served as a reference station throughout the robots' experiments, enabling relative differential altitude computation using \autoref{equation_alt_hyps}.
All platforms, except the altimeter rig, are equipped with a 16-beam lidar, MTi10 or Vectornav VN100 \acp{IMU}, and a DPS310 barometric pressure sensor. 
The platforms were deployed on the Athabasca Glacier, depicted in \autoref{fig:fig-intro}, and on the Université Laval campus.  

\subsection{\ac{MEMS} temperature calibration}
This section presents the results of calibrating \ac{MEMS}-based pressure sensors through temperature variation during static tests.
As every \ac{MEMS} sensor differs, the generic compensation in \autoref{equation_calibration} must be evaluated for each DPS310 sensor used.
We used the ceramic-based model 278 pressure transducer from Setra as the ground-truth since its measurement is not affected by temperature, unlike \ac{MEMS}-based sensors.
In order to evaluate the influence of the coefficients, we optimize five variations of the generic compensation equation from \ref{equation_calibration}, as
\begin{equation}
\begin{aligned}
\bm{A}_p &= \left[ 
\renewcommand\arraystretch{0.4}
\begin{array}{@{}c@{}c@{}c@{}c@{}}
\bullet & \circ & \circ & \circ \\
\bullet & \circ & \circ & \circ \\
\bullet & \circ & \circ & \circ \\
\bullet & \circ & \circ & \circ \\
\end{array}
\right]
& \bm{A}_\text{simple} =& \left[ 
\renewcommand\arraystretch{0.4}
\begin{array}{@{}c@{}c@{}c@{}c@{}}
\bullet & \bullet & \circ & \circ \\
\bullet & \circ & \circ & \circ \\
\bullet & \circ & \circ & \circ \\
\bullet & \circ & \circ & \circ \\
\end{array}
\right]
& \bm{A}_\text{ind} =& \left[ 
\renewcommand\arraystretch{0.4}
\begin{array}{@{}c@{}c@{}c@{}c@{}}
\bullet & \bullet & \bullet & \bullet \\
\bullet & \circ & \circ & \circ \\
\bullet & \circ & \circ & \circ \\
\bullet & \circ & \circ & \circ \\
\end{array}
\right]
\\
\bm{A}_\text{m} &= \left[ 
\renewcommand\arraystretch{0.4}
\begin{array}{@{}c@{}c@{}c@{}c@{}}
\bullet & \bullet & \circ & \circ \\
\bullet & \bullet & \circ & \circ \\
\bullet & \bullet & \circ & \circ \\
\bullet & \circ & \circ & \circ \\
\end{array}
\right]
& \bm{A}'_\text{m} =& \left[ 
\renewcommand\arraystretch{0.4}
\begin{array}{@{}c@{}c@{}c@{}c@{}}
\star & \star & \circ & \circ \\
\star & \star & \circ & \circ \\
\star & \star & \circ & \circ \\
\star & \circ & \circ & \circ \\
\end{array}
\right]
& \bm{A}_\text{full} =& \left[ 
\renewcommand\arraystretch{0.4}
\begin{array}{@{}c@{}c@{}c@{}c@{}}
\bullet & \bullet & \bullet & \bullet \\
\bullet & \bullet & \bullet & \bullet \\
\bullet & \bullet & \bullet & \bullet \\
\bullet & \bullet & \bullet & \bullet \\
\end{array}
\right] ,
\label{equation_A_matrices}
\end{aligned}
\end{equation}
where $[\bullet]$ are coefficients to be optimized, $[\circ]$ represents zeros, and $[\star]$ are the coefficients provided by the manufacturer.
First, the matrix $\bm{A}_p$ is intended to validate whether temperature significantly impacts the measurement.
Second, $\bm{A}_{\text{simple}}$ and $\bm{A}_{\text{ind}}$ both investigate to what extent temperature affects the measurement.
Then, $\bm{A}_m^{'}$ and $\bm{A}_m$ both represent the manufacturer-given compensation equation, but with $\bm{A}_m$ re-optimizing the parameters for this specific sensor.
Finally, $\bm{A}_{\text{full}}$ intends to take full advantage of the available coefficients to reach a lower error.

\begin{figure}[htbp]
	\centering
	\includegraphics[width=\linewidth]{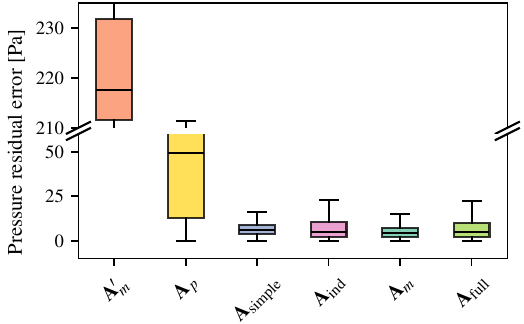}
	\caption{Residual error in Pascals obtained after minimizing the temperature compensation calibration models presented in \autoref{equation_A_matrices}. (Boxes and whiskers are truncated to help visualization)}
	\label{fig:sensor_calibration}
    \vspace{-2mm}
\end{figure}

To optimize the models, we conducted static measurements while heating the altimeter rig to \SI{50}{\celsius} and letting it cool down to \SI{5}{\celsius} in a refrigerated environment.
Positioning both the reference and the evaluated sensors at the same level and fixed next to each other, as shown in \autoref{fig:fig-intro}, ensured that the pressure measurements from both sensors should be identical. 
To collect the data, we heated the altimeter rig using a heat gun for \SI{45}{\minute} to an hour, until we reached a temperature of \SI{50}{\celsius}; then, we set the rig in a refrigerator to cool down to a temperature of about \SI{15}{\celsius}, which took approximately two hours. 
We repeated this process eight times and conducted the experiment using a freezer once, reaching as low as \SI{0}{\celsius}.
To complete the dataset, we added five static experiments at room temperature lasting one to two hours each to give this temperature zone more importance in the optimization as it is the main use range.
\autoref{fig:sensor_calibration} shows the residual pressure error between the pressure recorded with the Setra ceramic sensor and $p_\text{cal}$ for the five models optimized for calibration, as well as the factory calibration.
First, the factory calibration provided is not suited for temperature variations.
Then, as expected, there is a clear correlation between the temperature and pressure measurements, as $\bm{A}_p$ performs the worst out of the optimized models.
As for the other models, we see that as soon as we take into account the temperature, we reach similar results with a median residual error of \SI{6.28}{\pascal} and a standard deviation of \SI{4.71}{\pascal} for the $\bm{A}_\text{simple}$.
Therefore the model provided by the manufacturer relies on more parameters than necessary.
Thus, our final model is
\begin{align}
p_\text{cal} = c_{00} + c_{10} \, p_\text{raw} + c_{20} \, p_\text{raw}^2 + c_{30}\, p_\text{raw}^3 + c_{01} \, t,
\end{align}
for which we do not provide the coefficients, as they are specific to each individual sensor.

\subsection{Sensitivity analysis}
\begin{figure}[tbp]
	\centering
	\includegraphics[width=\linewidth]{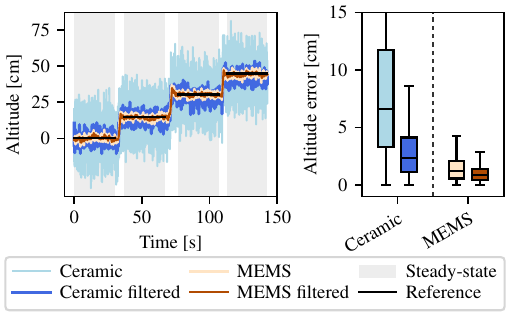}
	\caption{
 Uncertainty results of ceramic and \ac{MEMS} pressure sensor converted to altitude measurements.
 \emph{Left}: estimations of relative altitude through time for steps of \SI{13}{cm} (black) with an unfiltered ceramic sensor (light blue), a filtered ceramic sensor (blue), and two MEMS pressure signals unfiltered (beige) and filtered (brown). The steady-state (grey) refers to the measurement window during which the sensors remain stationary at the specified fixed heights.
 \emph{Right}: errors in steady-state of the four altitude-estimation methods for smaller steps of \SI{3}{cm} with unfiltered and filtered signals.
 }
	\label{fig:sensitivity_analysis}
    \vspace{-2mm}
\end{figure}

To reach centimeter-level accuracy, we must be able to detect pressure variations of \SI{0.1}{\pascal}.
Therefore, this section presents the sensitivity analysis performed to evaluate the noise level of both the \ac{MEMS}-based and ceramic-based sensors.
The experiments consisted of moving the altimeter rig up and down at known relative heights to evaluate the precision of the sensor technology.
Another DPS310 \ac{MEMS}-based sensor was used as the base station to compute the relative altitude.
First, a qualitative experiment with steps of \SI{13}{cm} was conducted to evaluate the quality of the signals, as shown on the left in \autoref{fig:sensitivity_analysis}.
Second, a quantitative analysis was conducted over an experiment where the altimeter rig was moved up and down a step of \SI{3}{cm} 10 times.
We evaluated the steady-state error on the altitude for unfiltered signals and filtered signals in \autoref{fig:sensitivity_analysis} (right).

With the qualitative experiment (\autoref{fig:sensitivity_analysis} (left)), there is a clear highlight that filtering is necessary for the ceramic-based sensor to achieve performances suitable for localization solutions.
The filtered signals were obtained using a \acf{KF}, the parameters of which were tuned using the signal's noise characteristics.
In the quantitative analysis (\autoref{fig:sensitivity_analysis} (right)), a median error of \SI{2.36}{cm} and a standard deviation of \SI{2.25}{cm} can be achieved through this filtering of the ceramic sensor.
\autoref{fig:sensitivity_analysis} (right) also shows we achieve a median error of \SI{0.834}{cm} and a standard deviation of \SI{0.725}{cm}, thus reaching a slightly more biased solution with three times less noise on the measurement than the ceramic sensor. 
Thus, we chose to use the DPS310 \ac{MEMS}-based barometric pressure sensor to move forward with our \ac{ICP} implementation since a lower error level was attainable and since the bias can be ignored through a locally consistent mapping framework.
Even though more calibration and bias mitigation have to be done for the \ac{MEMS} sensor, all of these are possible within the context of mobile robotics.
Stronger performances could also be attained with the ceramic-based sensor, using a more aggressive filtering, which would avoid the temperature calibration of \ac{MEMS} while also including a delay in the signal.

\subsection{Increased localization performances}
We conducted an experiment within a building to assess the performance of \ac{ICP} enhanced with pressure measurements.
To achieve a valid evaluation, we drove the robot down three floors with known heights, therefore exciting the z-coordinate over \SI{12}{\meter} and for a path length of \SI{311}{m}, excluding the staircases.
Thus, distances of approximately \SI{100}{\metre} were traveled at fixed heights on three different floors of a building (\ie in corridors), limiting the relative z-coordinate of the map to the height of each level and emphasizing the impact of our implementation on degenerate environments.
This experiment allows us to compare the performance of localization algorithms using more data than a typical loop, where only the start and end points could be used to evaluate the results.
We compared our solution against two other algorithms, namely the 6\nobreakdashes-\ac{DOF} \ac{ICP}, optimizing the full pose of the robot for every scan registration~\citep{Pomerleau2013}, and the 4\nobreakdashes-\ac{DOF} \ac{ICP}, constraining the roll and pitch angles using the gravity vector and optimizing the remaining four \acp{DOF}~\citep{Kubelka2022}.
Both of these algorithms were provided with the same prior transformation computed from a fusion of \ac{IMU} and wheel encoder measurements.
Finally, we compared the gain of simply adding our altitude measurement fused with the \ac{IMU} and wheel encoders as prior transformation, named \emph{X\nobreakdashes-\ac{DOF} altitude} in the results.

\begin{figure}[t]
	\centering
	\includegraphics[trim={13px 0 0 0}, width=\linewidth]{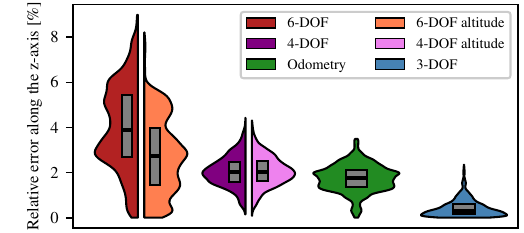}
\caption{Relative localization error on the z-coordinate computed with an adapted \ac{RPE}, using a computation distance of \SI{5}{m}, from a mapping experiment on three floors of a building. Showcasing 6\nobreakdashes-\ac{DOF}, 4\nobreakdashes-\ac{DOF}, and 3\nobreakdashes-\ac{DOF} implementations of \ac{ICP}, 6\nobreakdashes-{DOF} altitude and 4\nobreakdashes-\ac{DOF} altitude \ac{ICP} implementation which include the altitude measurement in their prior transformation, and the wheel-\ac{IMU} odometry.}
	\label{fig:Loc_error}
    \vspace{-2mm}
\end{figure}

\autoref{fig:Loc_error} depicts the relative error computed on the z-coordinate of the trajectory, as it is the main drift targeted with our implementation.
To compute the relative error, we evaluate the drift of the z-coordinate over distances of \SI{5}{\meter} using an adapted \ac{RPE} metric, and remove the staircases from the trajectory, as we only had access to each floor height as ground truth.
We can observe a significant improvement in the z-coordinate localization error of 3\nobreakdashes-\ac{DOF} over both other \ac{ICP} implementations and over wheel-\ac{IMU} odometry.
Also, we can see that this odometry solution performs on par with 4\nobreakdashes-\ac{DOF}, which is easily explained by the fact that our experiment was conducted on planar floors with little vibration, and the stairs were removed from the evaluation.
In this particular experiment, we observe the 3\nobreakdashes-\ac{DOF} improving the drift in elevation by \SI{84}{\percent}, with the median \ac{RPE} of 4\nobreakdashes-\ac{DOF} being \SI{2.02}{\percent} compared to the 3\nobreakdashes-\ac{DOF} being \SI{0.31}{\percent}.
Moreover, \autoref{fig:Loc_error} shows that adding the altitude as prior information can benefit 6\nobreakdashes-\ac{DOF} by improving its performance by \SI{30}{\percent} while not hindering the performance of 4\nobreakdashes-\ac{DOF}.

\begin{figure}[t]
    \centering
    \includegraphics[width=\linewidth]{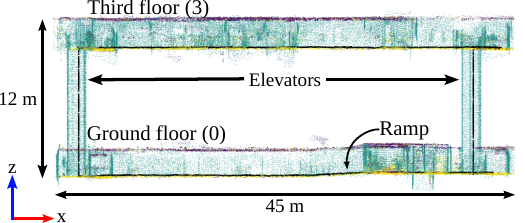}
    \caption{Map of the third floor and ground floor of the university building showing the accuracy and performance of our 3\nobreakdashes-\ac{DOF} \ac{ICP}. It can be seen that the small ramp of the ground floor is detected and that the map reaches the correct elevation on the way back up to the third floor, all through the use of elevators.}
    \vspace{-2mm}
    \label{fig:icp_map}
\end{figure}

We can qualitatively observe how the 3D map maintains its global consistency in elevation in \autoref{fig:icp_map}.
The figure shows the resulting map of a second experiment where the building's third floor and ground floor were driven, using elevators to move between the two floors.
The two floors of the building can easily be distinguished and also correspond within centimeters to our manual measurement of each floor's height.
Furthermore, \autoref{fig:icp_map} shows how constraining the altitude for the \ac{ICP} minimization enables the use of elevators.

\vspace{-1mm}
\section{Conclusion}
\label{sec:conclusion}
\vspace{-1mm}
In this paper, we proposed a novel 3\nobreakdashes-\ac{DOF} \ac{ICP} algorithm based on barometer constraints.
Through a thorough evaluation and calibration of barometers, we showed that \ac{MEMS} sensors are better fitted for robotics applications, reaching centimeter-level accuracy with lower filtering requirements.
Although further affected by noise, ceramic sensors could also be used with advanced signal filtering, allowing for the avoidance of bias caused by the internal temperature of the \ac{MEMS} sensor.
Using the proposed novel calibration, the altimeter measurements were used to constrain the \ac{ICP} algorithm, reaching an \ac{RPE} on the vertical axis of \SI{0.31}{\percent}.
Future works will focus on implementing penalties on the \ac{ICP} algorithm rather than fully constraining the vertical axis, thus also leveraging structural information from the environment, such as floors and ceilings.
Further evaluation of the algorithm with outdoor large-scale experiments will be necessary to push the boundaries of our implementation, along with comparisons to other state-of-the-art algorithms.

\vspace{-1mm}
\section*{Acknowledgment}
\vspace{-1mm}
This research was supported by the  Natural  Sciences and Engineering  Research  Council of  Canada  (NSERC)  through the grant CRDPJ 527642-18 SNOW (Self-driving Navigation Optimized for Winter).


\IEEEtriggeratref{6}
\IEEEtriggercmd{\enlargethispage{-0.1in}}

\printbibliography

@inproceedings{Song2015,
  title={Towards autonomous control of quadrotor unmanned aerial vehicles in a {GPS}-denied urban area via laser ranger finder},
  author={Song, Yinglin and Xian, Bin and Zhang, Yao and Jiang, Xinran and Zhang, Xu},
  booktitle={Optik},
  volume={126},
  number={23},
  pages={3877--3882},
  year={2015},
  publisher={Elsevier}
}

@inproceedings{Urzua2019,
  title={Monocular {SLAM} system for {MAV}s aided with altitude and range measurements: {A} {GPS}-free approach},
  author={Urzua, Sarquis and Mungu{\'\i}a, Rodrigo and Grau, Antoni},
  booktitle={Journal of Intelligent \& Robotic Systems},
  volume={94},
  pages={203--217},
  year={2019},
  publisher={Springer}
}

@inproceedings{Talbot2023,
  title={Principled {ICP} {C}ovariance {M}odelling in {P}erceptually {D}egraded {E}nvironments for the {EELS} {M}ission {C}oncept},
  author={Talbot, William and Nash, Jeremy and Paton, Michael and Ambrose, Eric and Metz, Brandon and Thakker, Rohan and Etheredge, Rachel and Ono, Masahiro and Ila, Viorela},
  booktitle={2023 IEEE/RSJ International Conference on Intelligent Robots and Systems (IROS)},
  pages={10763--10770},
  year={2023},
  organization={IEEE}
}

@inproceedings{Zaliva2014,
  title={Barometric and {GPS} altitude sensor fusion},
  author={Zaliva, Vadim and Franchetti, Franz},
  booktitle={2014 IEEE International Conference on Acoustics, Speech and Signal Processing (ICASSP)},
  pages={7525--7529},
  year={2014},
  organization={IEEE}
}

@inproceedings{Bolanakis2017,
  title={{MEMS} barometers and barometric altimeters in industrial, medical, aerospace, and consumer applications},
  author={Bolanakis, Dimosthenis E},
  booktitle={IEEE Instrumentation \& Measurement Magazine},
  volume={20},
  number={6},
  pages={30--55},
  year={2017},
  publisher={IEEE}
}

@inproceedings{parviainen2008differential,
  title={Differential barometry in personal navigation},
  author={Parviainen, Jussi and Kantola, Jussi and Collin, J},
  booktitle={2008 IEEE/ION Position, Location and Navigation Symposium},
  pages={148--152},
  year={2008},
  organization={IEEE}
}

@inproceedings{Sabatini2013,
  title={A stochastic approach to noise modeling for barometric altimeters},
  author={Sabatini, Angelo Maria and Genovese, Vincenzo},
  booktitle={Sensors},
  volume={13},
  number={11},
  pages={15692--15707},
  year={2013},
  publisher={Molecular Diversity Preservation International (MDPI)}
}

@inproceedings{Matyja2022,
  title={The {MEMS}-based barometric altimeter inaccuracy and drift phenomenon},
  author={Matyja, Tomasz and Kubik, Andrzej and Stanik, Zbigniew},
  booktitle={Scientific Journal of Silesian University of Technology. Series Transport},
  year={2022}
}

@inproceedings{Babin2019,
  title = {Large-scale 3{D} {M}apping of {S}ubarctic {F}orests},
  booktitle = {Proceedings of the Conference on Field and Service Robotics (FSR). Springer Tracts in Advanced Robotics},
  year = {2019},
  author = {Babin, P. and Dandurand, P. and Kubelka, V. and Gigu{\`e}re, P. and Pomerleau, F.},
  project = {penalty_icp,libpointmatcher},
}

@inproceedings{Kubelka2022,
  title = {Gravity-constrained point cloud registration},
  author = {Kubelka, Vladimír and Vaidis, Maxime and Pomerleau, François},
  booktitle = {Proceedings of the IEEE International Conference on Intelligent Robots and Systems (IROS)},
  doi = {10.1109/IROS47612.2022.9981916},
  year = {2022},
  pages = {4873--4879},
}

@article{Ebadi2023,
  author={Ebadi, Kamak and Bernreiter, Lukas and Biggie, Harel and Catt, Gavin and Chang, Yun and Chatterjee, Arghya and Denniston, Christopher E. and Deschênes, Simon-Pierre and Harlow, Kyle and Khattak, Shehryar and Nogueira, Lucas and Palieri, Matteo and Petráček, Pavel and Petrlík, Matěj and Reinke, Andrzej and Krátký, Vít and Zhao, Shibo and Agha-mohammadi, Ali-akbar and Alexis, Kostas and Heckman, Christoffer and Khosoussi, Kasra and Kottege, Navinda and Morrell, Benjamin and Hutter, Marco and Pauling, Fred and Pomerleau, François and Saska, Martin and Scherer, Sebastian and Siegwart, Roland and Williams, Jason L. and Carlone, Luca},
  journal={IEEE Transactions on Robotics}, 
  title={Present and Future of SLAM in Extreme Environments: The DARPA SubT Challenge}, 
  year={2024},
  volume={40},
  number={},
  pages={936-959},
  doi={10.1109/TRO.2023.3323938}}

@inproceedings{Leisiazar2022,
  title={Real-time {M}apping of {M}ulti-{F}loor {B}uildings {U}sing {E}levators},
  author={Leisiazar, Sahar and Mahdavian, Mohammad and Park, Edward J and Chen, Mo},
  booktitle={2022 IEEE/ASME International Conference on Advanced Intelligent Mechatronics (AIM)},
  pages={314--321},
  year={2022},
  organization={IEEE}
}

@article{Dellaert2012,
  title={Factor graphs and {GTSAM}: {A} hands-on introduction},
  author={Dellaert, Frank},
  journal={Georgia Institute of Technology, Tech. Rep},
  volume={2},
  pages={4},
  year={2012}
}

@article{Pomerleau2015,
  title={A review of point cloud registration algorithms for mobile robotics},
  author={Pomerleau, Fran{\c{c}}ois and Colas, Francis and Siegwart, Roland and others},
  journal={Foundations and Trends{\textregistered} in Robotics},
  volume={4},
  number={1},
  pages={1--104},
  year={2015},
  publisher={Now Publishers, Inc.}
}

@inproceedings{Hipkin2000,
  title={Modelling the geoid and sea-surface topography in coastal areas},
  author={Hipkin, Roger},
  booktitle={Physics and Chemistry of the Earth, Part A: Solid Earth and Geodesy},
  volume={25},
  number={1},
  pages={9--16},
  year={2000},
  publisher={Elsevier}
}

@article{Baril2022,
   author = {Dominic Baril and Simon-Pierre Deschênes and Olivier Gamache and Maxime Vaidis and Damien LaRocque and Johann Laconte and Vladimír Kubelka and Philippe Giguère and François Pomerleau},
   doi = {10.55417/fr.2022050},
   issue = {1},
   journal = {Field Robotics},
   pages = {1628-1660},
   title = {Kilometer-scale autonomous navigation in subarctic forests: challenges and lessons learned},
   volume = {2},
   url = {https://doi.org/10.55417/fr.2022050},
   year = {2022},
}

@book{wallace2006atmospheric,
  title={Atmospheric science: an introductory survey},
  author={Wallace, John M and Hobbs, Peter V},
  volume={92},
  year={2006},
  publisher={Elsevier}
}

@book{Barfoot2017,
  title={State {E}stimation for {R}obotics},
  author={Barfoot, T.D.},
  isbn={9781107159396},
  lccn={2017010237},
  year={2017},
  publisher={Cambridge, UK: Cambridge University Press}
}

@inproceedings{dubois2024workshop,
      title={3{D} {M}apping of {G}lacier {M}oulins: {C}hallenges and lessons learned}, 
      author={William Dubois and Matěj Boxan and Johann Laconte and François Pomerleau},
      booktitle={2024 International Conference on Robotics and Automation (ICRA) Workshop on Field Robotics},
      year={2024},
      eprint={2404.18790},
      archivePrefix={arXiv},
      primaryClass={cs.RO},
      url={https://arxiv.org/abs/2404.18790}, 
}

@article{Pomerleau2013,
  title = {Comparing {ICP} variants on real-world data sets: Open-source library and experimental protocol},
  journal = {Autonomous Robots},
  year = {2013},
  volume = {34},
  number = {3},
  pages = {133-148},
  author = {Pomerleau, F. and Colas, F. and Siegwart, R. and Magnenat, S.},
  project = {libpointmatcher, scutigera}
}
\end{document}